\documentclass[conference]{IEEEtran}
\IEEEoverridecommandlockouts
\usepackage{setspace}
\usepackage{cite}
\usepackage{amsmath,amssymb,amsfonts}
\usepackage{graphicx}
\usepackage{textcomp}
\usepackage{xcolor}
\usepackage{amsthm}
\usepackage{algorithm}
\usepackage{algpseudocode}

\usepackage{bm} % Required for bold vectors (\bm command)
\DeclareMathOperator*{\argmin}{argmin}
\def\BibTeX{{\rm B\kern-.05em{\sc i\kern-.025em b}\kern-.08em
    T\kern-.1667em\lower.7ex\hbox{E}\kern-.125emX}}

\begin{document}

\title{Self-Supervised Learning-Based Path Planning and Obstacle Avoidance Using PPO and B-Splines in Unknown Environments\\}
%{\footnotesize \textsuperscript{*}Note: Sub-titles are not captured in %Xplore and

\author{\IEEEauthorblockN{Shahab Shokouhi}
\IEEEauthorblockA{\textit{Department of Mechanical Engineering} \\
\textit{Univerisity of New Hampshire} \\
Durham, New Hampshire, USA \\
shahaab.shokouhi@unh.edu}

\and
\IEEEauthorblockN{May-Win Thein}
\IEEEauthorblockA{\textit{Department of Mechanical Engineering} \\
\textit{Univerisity of New Hampshire}\\
Durham, New Hampshire, USA \\
may-win.thein@unh.edu}

\and
\IEEEauthorblockN{Oguzhan Oruc}
\IEEEauthorblockA{\textit{Department of Mechanical Engineering} \\
\textit{The Citadel School of Engineering}\\
Charleston County, South Carolina, USA \\
ooruc@citadel.edu}
}

\maketitle

\begin{abstract}
This paper introduces Smart B-Splines (SmartBSP), an advanced self-supervised learning framework for real-time path planning and obstacle avoidance in autonomous robotics navigating through complex environments. The proposed system integrates Proximal Policy Optimization (PPO) with Convolutional Neural Networks (CNN) and Actor-Critic architecture to process limited LiDAR inputs and compute spatial decision-making probabilities. The robotic vehicle's perceptual field is discretized into a grid format, which the CNN analyzes to produce a spatial probability distribution. During the training process a nuanced cost function is minimized that accounts for path curvature, endpoint proximity, and obstacle avoidance. Simulations results in different scenarios validate the algorithm's resilience and adaptability across diverse operational scenarios. Subsequently, Real-time experiments, employing the Robot Operating System (ROS), are carried out to assess the efficacy of the proposed algorithm. The experimental results show that the proposed algorithm is successful in different scenarios while keeping the path as smooth as possible. 
\end{abstract}


\section{Introduction}
Autonomous navigation and environment mapping for autonomous vehicles has garnered considerable attention in the field of robotics and autonomous systems research \cite{b1}. More specifically, path planning methodologies are divided into two primary categories: (1) those focusing on navigation within entirely unknown environments \cite{b2}, with only access to real-time range sensor data, and (2) those dedicated to environments that are either partially or completely known \cite{b3}. Classic foundations spanning grid- and sampling-based planning further contextualize these categories, including A* and heuristic search, probabilistic roadmaps, rapidly-exploring random trees and optimal variants \cite{b19,b20,b21,b22}. The interest in path planning in unknown environments is driven by the critical need for autonomous vehicles to operate effectively in situations where prior knowledge of the environment is absent or severely limited. Examples of such applications include extraterrestrial rover missions on planets like Mars, where the terrain is largely uncharted, and search-and-rescue operations in disaster-stricken areas, where rapid changes in the environment preclude the use of pre-existing maps. In such settings, classical local strategies (e.g., Bug algorithms and frontier-based exploration) have long provided baselines \cite{b23,b24}.

Additionally, the advent of Artificial Intelligence (AI) has revolutionized the way researchers tackle the path planning problem \cite{b4}, \cite{b5}. By harnessing the power of AI, scholars are now able to construct neural networks that not only learn from simulated and real data but also exhibit human-like expertise in decision-making processes \cite{b6}. Beyond supervised pipelines, differentiable planning modules and learned planners such as Value Iteration Networks have shown promise in embedding planning structure into deep models \cite{b28}. However, the lack of access to useful data makes the offline training of neural networks challenging. In such situations, self-supervised learning methods have proven to be effective \cite{b7}. However, in situ real-time learning during field operations can be time consuming and sometimes unsafe \cite{b8}. To ensure safe learning, Proximal Policy Optimization (PPO) \cite{b16} has been employed for offline path planner training. While PPO is widely known as a method in Reinforcement Learning (RL), this study assumes that there is no transition between states (i.e., observations), that is, the impact of decisions made at any time step on future events is ignored.

While the safety of the path plays a major role in the path planning task, the smoothness of the path is equally important. Due to the low maneuverability of underactuated vehicles, paths with high curvature or sudden changes in heading are not feasible. In recent years, B-splines have been employed in numerous studies to minimize the curvature of the path \cite{b9,b10,b11}. Related trajectory-optimization approaches such as CHOMP and STOMP also explicitly shape smoothness and collision costs for feasible motion generation \cite{b26,b25}. While these studies have made notable advancements in incorporating curvature and smoothness into path planning, the computational demand and possible numerical instability of their methods—especially when dealing with non-convex obstacles—present significant challenges for real-time implementation. 

The SmartBSP method proposed in this paper incrementally builds available path via LiDAR-based sensor feedback. The main objective of the proposed algorithm is to determine  optimal positions of control points for local B-splines that guide the vehicle toward the target while avoiding obstacles and minimizing curvature.
The proposed path planning algorithm is:
\begin{itemize}
    \item tailored for completely unknown environments, capable of overcoming various non-convex obstacles.
    \item designed to explicitly incorporates path smoothness.
    \item designed for rapid execution, enabling real-time path planning.
    \item vehicle and environment-agnostic via hyperparameter adjustment (e.g., field of view and range).
    \item applicable to any generic target location and, therefore, does not require path planner retraining .
\end{itemize}

The remainder of the paper is structured as follows: Section II reviews related work, providing an overview of existing AI-based path planning methodologies and their limitations. Section III delves into the methodology, detailing the various components involved in the SmartBSP algorithm, including the integration of Proximal Policy Optimization (PPO) with Convolutional Neural Networks (CNNs) and the Actor-Critic architecture. Section IV discusses the simulation results, highlighting the performance of the SmartBSP algorithm across different scenarios and the parameters used in the simulations. Section V focuses on the experimental results, showcasing the practical implementation and testing of the SmartBSP algorithm on a real-world robotic platform. The conclusion summarizes the findings and outlines potential directions for future research.

\section{Related Work}
In recent years, deep learning algorithms have been used in many studies to enhance traditional path planning methods. For example, in \cite{b12}, Convolutional Neural Networks (CNNs) are employed to learn pre-planned paths from other offline path planning methods, marking a significant step towards leveraging image-based methodologies in path planning. In \cite{b4}, Double Deep Q-Learning is utilized to navigate path planning challenges within dynamically changing environments.
Sharma et al. \cite{b5} have employed Q-learning techniques to determine the optimal path in a known environment, with the specific limitation that the goal's location is not generalized across different scenarios. In \cite{b13}, Deep Q-learning and CNNs are utilized to construct paths within grid patches derived from the environment. Wang et al. \cite{b14} leveraged neural networks to improve the probability distribution of sampling in Rapidly-exploring Random Trees Star ($RRT^*$) algorithm, aiming to accelerate the convergence towards an optimal solution. Pflueger et al. \cite{b7} proposed an approach using inverse reinforcement learning with soft value iteration networks (SVIN) for planetary rover path planning. This method leverages deep convolutional networks and value iteration networks to handle the complexities of navigation on Mars, demonstrating its effectiveness on both grid world and realistic Mars terrain datasets. Chiang et al. \cite{b15} introduced RL-RRT, a kinodynamic motion planner that combines reinforcement learning with sampling-based planning. Their method uses deep reinforcement learning to train an obstacle-avoiding policy, which serves as a local planner, and a reachability estimator that predicts the time to reach a state in the presence of obstacles. Complementary lines of work integrate differentiable planners (e.g., VIN) and learning–sampling hybrids (e.g., PRM-RL) for long-range navigation \cite{b28,b30}. For reactive multi-agent avoidance and local navigation baselines, velocity-obstacle methods such as ORCA are widely used \cite{b31}. On the optimization side, CHOMP and STOMP remain influential for smooth, collision-aware motion generation \cite{b26,b25}. The next section describes the methodology and steps involved in our proposed method.

\section{Methodology}
The primary objective of the proposed path planning method is to determine the optimal control points of B-splines, similar to that of \cite{b9,self}, but diverges from the authors' previous works by integrating a novel approach that: 1) accelerates the optimization process in real-time and 2) it is capable of addressing many non-convex obstacles in an unknown environment. Central to this methodology is the consideration of both curvature and the spatial distance of the final point of each local B-spline to the target, while concurrently ensuring avoidance of detected obstacles within a given sensor range.

\begin{figure}[htp]
\centering
\includegraphics[width=\linewidth]{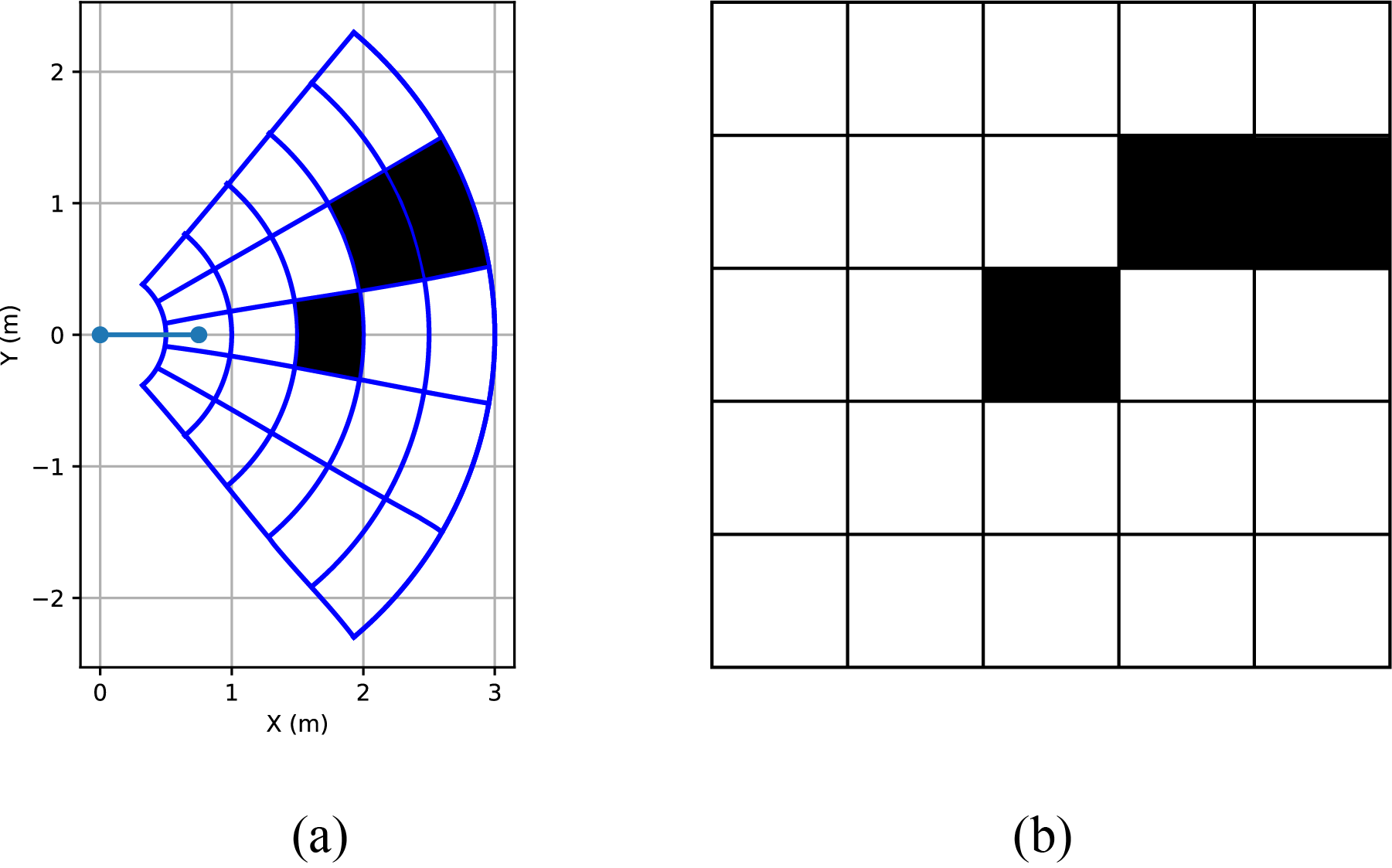}
\caption{(a) Circular grid formed based on observed point cloud, (b) equivalent square grid}
\label{fig1}
\end{figure}

Distinctive to this study is the initial transformation of the perceived point cloud into an $n\times n$ circular grid representation, which is depicted in Fig.~\ref{fig1}~(a). Grid cells that have more than $m$ points are considered as obstacles. In this paper, a $5\times 5$ grid configuration has been employed, comprising $5$ angular intervals and $5$ radial intervals. The transformation emulates the configuration of contemporary range sensors commonly deployed in autonomous vehicle platforms. This preprocessing step enhances the method's applicability, rendering it more congruent with real-world scenarios.

Subsequently, the circular grid undergoes a conversion process to a square grid format (Fig.~\ref{fig1}~(b)). This conversion facilitates seamless integration with a convolutional neural network (CNN), a pivotal component of the proposed methodology. The CNN operates on the square grid input and outputs the probability distribution governing the selection of each control point for the B-spline trajectory. Here, the angular coordinate of the circular grid corresponds to the $y$ coordinates in the square coordinate system, whereas the radial coordinate aligns with the $x$ coordinates. 

\subsection{Environment}
In the context of an self-supervised learning problem, the environment serves as a component that processes inputs or actions alongside current states, yielding subsequent rewards or costs. Formally, we define the environment as $\mathcal{E}=(C,\mathcal{S})$, where $\mathcal{S}$ represents the observed point cloud (state) and $C$ signifies the cost function, noting here that the choice between reward maximization and cost minimization is arbitrary, as maximizing reward can be equivalently framed as minimizing cost. It is assumed that there is no transition between states. In other words, the learning problem focuses on improving the outcome by adjusting the actions (i.e., control points of the B-spline) for each given $5\times 5$ grid configuration.

The radial coordinates of the control points are predetermined, comprising $n+1$ radial coordinates incrementally increasing from zero to the sensor range limit, inclusive of the origin which signifies the location of the robot at each time step. Since the location of the first and second control points is predetermined, as shown in Fig.~\ref{fig1}~(a), the learning algorithm is left to choose among $n-1$ distinct angular coordinates (actions) for the remaining control points. The first and second control points are chosen to be at fixed positions relative to the robot so as to mitigate abrupt changes in the desired heading angle. Therefore, the set of all control points (actions) $\mathcal{A}$ can be written as:
\begin{equation*}
    \mathcal{A} = \{\mathcal{A}_1, \mathcal{A}_2, ... , \mathcal{A}_{n+1}\}
\end{equation*}
Control points $\mathcal{A}_3,..., \mathcal{A}_{n+1}$ represent the geometric centers of the chosen cell grids. Given $\mathcal{A}$, the cost of the B-spline can be calculated as
\begin{equation}
C(\mathcal{S}, \mathcal{A}) = \rho_1 C_{\text{dist}} + \rho_2 C_{\text{curv}} + \rho_3 C_{\text{obs}}
\label{eq1}
\end{equation}

In this equation, $\rho_i$, where $i = 1, 2, 3$, are the weights, $C_{\text{dist}}$ represents the cost imposed due to the L2 norm between the final point of the B-spline and the final target location such that
\begin{equation}
    C_{\text{dist}}(\mathcal{A}) = \|\mathcal{A}_{n+1} - \mathcal{G}_{\text{final}}\|_2 
\end{equation}
where $\mathcal{G}_{\text{final}}$ is the target location. The term $C_{\text{curv}}$ denotes the cost imposed by the overall curvature of the B-spline, and $C_{\text{obs}}$ denotes the cost imposed due to collision with obstacles. In this paper, we define the cost imposed by the overall curvature of B-spline as
\begin{equation}
C_{\text{curv}}(\mathcal{A}) = \int_{\mathcal{A}_1}^{\mathcal{A}_{n+1}} k(\beta)^2 \, d\beta
\end{equation}
where $\beta$ is any point on the B-spline and $k(\beta)$ is defined at each point as
\begin{equation}
k(\beta) = \frac{y''(\beta)}{(1+y'(\beta)^2)^{3/2}}
\label{curvature}
\end{equation}
noting that all the derivatives in (\ref{curvature}) represent partial derivatives with respect to $\beta$. Furthermore, $C_{\text{obs}}$ is
\begin{equation}
    \begin{minipage}{0.9\linewidth}
    \centering
    $C_{\text{obs}}(\mathcal{S},\mathcal{A})= 
        \begin{cases}
            1 & \text{if } \text{collision} = \text{TRUE}\\
            0 & \text{otherwise} 
        \end{cases}$ \\
    \end{minipage}
\end{equation}

\subsection{Proximal Policy Optimization}
The framework of the self-supervised learning problem is such that $(\mathcal{S}, \mathcal{A}, C)$ is defined where $\mathcal{S}$ denotes the current state, representing the observed circle segment captured by the range sensor; $\mathcal{A} = (\mathcal{A}_{1},..., \mathcal{A}_{n+1})$ represents the set of $n+1$ actions (i.e., control points) selected at the current state; $C$ signifies the cost associated with choosing $\mathcal{A}$ at $\mathcal{S}$, calculated via (\ref{eq1}). Note that each $\mathcal{A}_{i}$, $i=3,...,n+1$, is sampled from a discrete distribution over $n$ possible outcomes.

PPO \cite{b16} can be characterized as a learning method that learns by increasing the probability of selecting actions that outperform the average. The primary distinction between PPO and other reinforcement learning methods is that it restricts the size of the steps taken in gradient descent to mitigate the risk of local optima. PPO incorporates the Clipped Surrogate Objective (CSO) and the utilization of multiple epochs of stochastic gradient ascent for each policy update. Here, the policy is defined as
\begin{equation}
\pi_{\theta}(a|\mathcal{S}) = P(\mathcal{A} = a|\mathcal{S})
\end{equation}
where $\theta$ are the parameters of the neural network and $a$ is the sampled action vector. Since each B-spline consists of $n-1$ different actions in addition to the fixed first and second points, the probability of choosing each specific B-spline or path is as follows:
\begin{equation}
    \pi_{\theta}(a|\mathcal{S}) = \prod_{i=3}^{n+1} \pi_{\theta}(a_{i}|\mathcal{S})
\end{equation}
or, alternatively in the $log$ form
\begin{equation}
    \log \pi_{\theta}(a|\mathcal{S}) = \sum_{i=3}^{n+1} \log \pi_{\theta}(a_{i}|\mathcal{S})
\end{equation}

PPO utilizes the ratio of the probability to constrain the policy update. The ratio is defined as
\begin{equation*}
    r(\theta) = \frac{\pi(a|\mathcal{S})}{\pi_{\text{old}}(a|\mathcal{S})}
\end{equation*}
This is achieved through the CSO function:
\begin{equation}
    L^{\text{clip}} (\theta) = \hat{\mathbb{E}}[\min{(r(\theta) \hat{A},      \\clip(r(\theta), 1-\epsilon, 1+\epsilon)\hat{A})}]
    \label{eq8}
\end{equation}
Here, the hyperparameter $\epsilon$ is typically set to a specific value, for instance, $\epsilon = 0.2$ \cite{b16}, and $\hat{A}$ represents the advantage function, distinct from the action vector $\mathcal{A}$:
\begin{equation}
    \hat{A} = C(\mathcal{S}, a) - \mathbb{E}_{a \in \mathcal{A}}[C(\mathcal{S}, a)]
\end{equation}

\subsection{Actor-Critic Architecture}
\begin{figure}[htp]
\centering
\includegraphics[width=\linewidth]{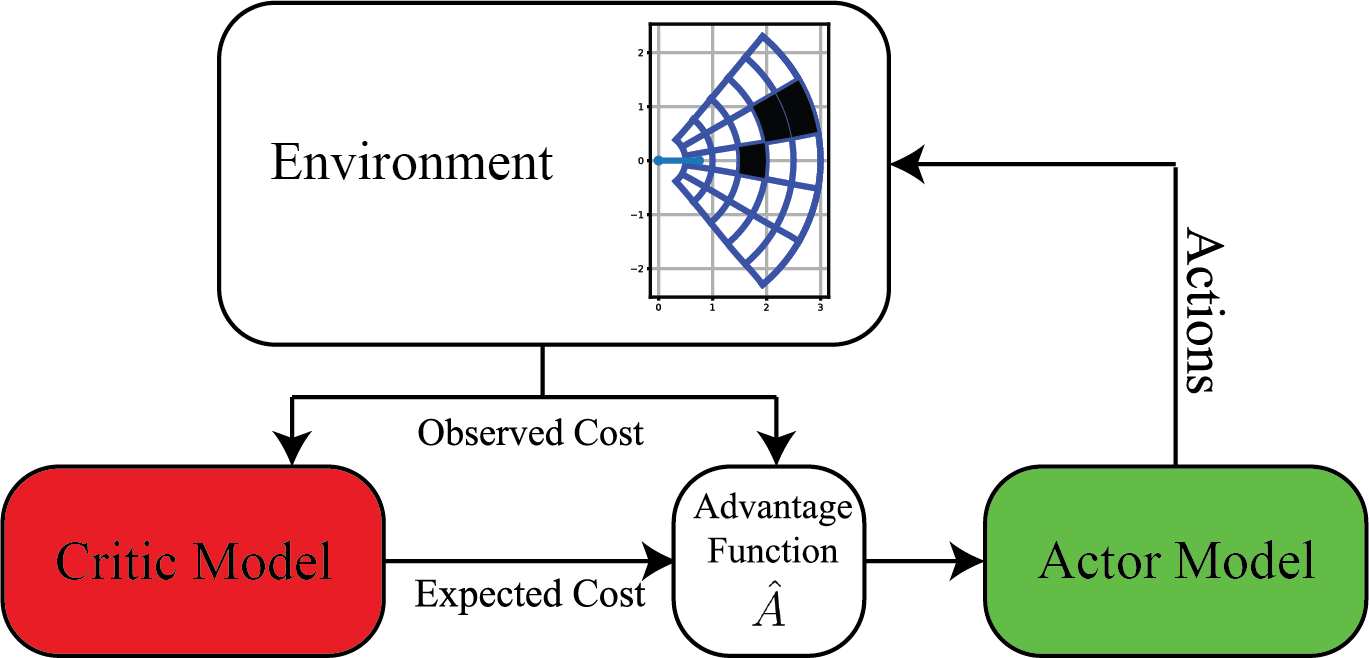}
\caption{Actor-Critic architecture}
\label{fig2}
\end{figure}

The dual-component setup of actor-critic architecture enables it to efficiently learn complex policies in environments with high-dimensional state and action spaces, making it well-suited for a wide range of learning tasks. The overall configuration of the AC architecture used in this paper is shown in Fig.~\ref{fig2}.

\begin{figure}[htp]
\centering
\includegraphics[width=\linewidth]{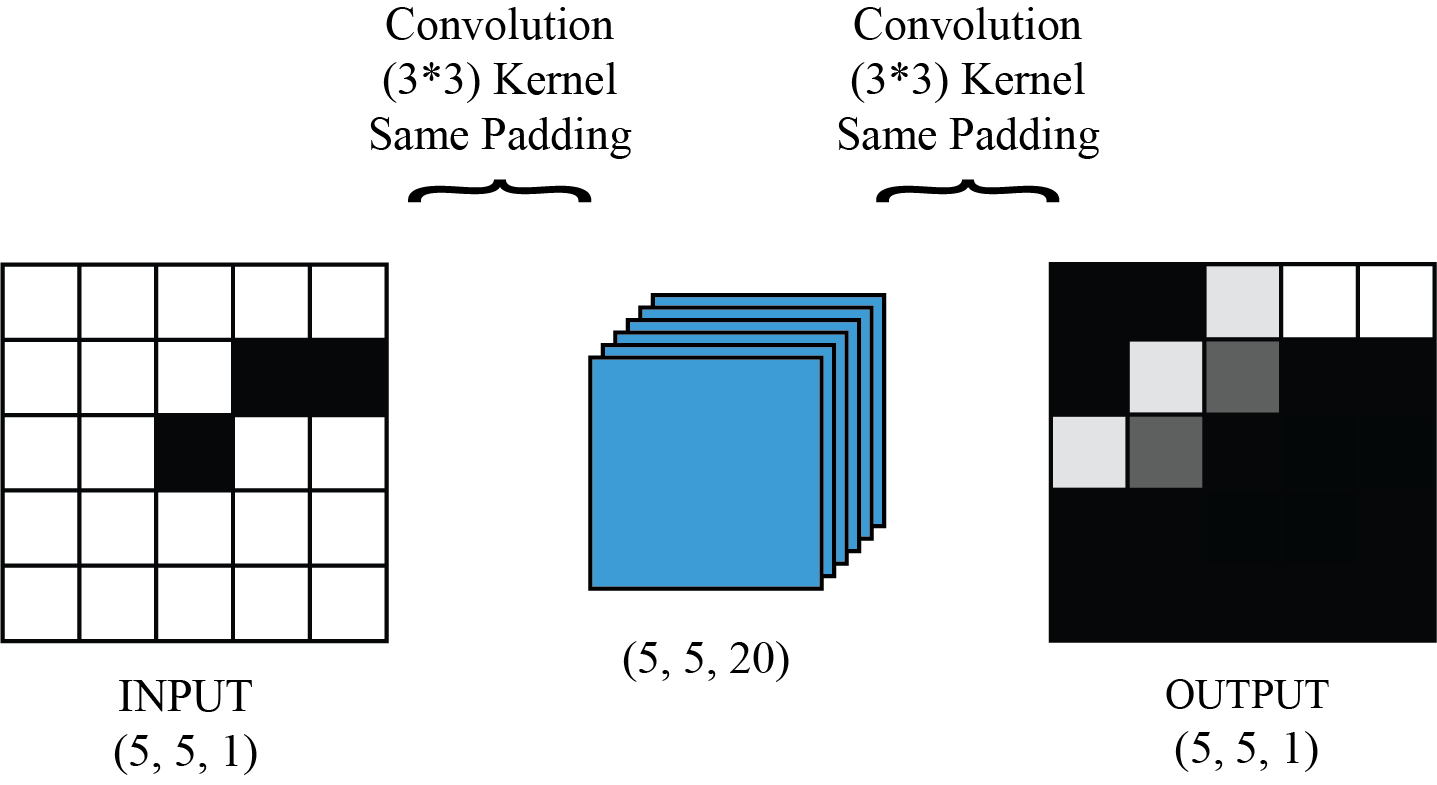}
\caption{Actor network architecture}
\label{fig3}
\end{figure}

\begin{figure}[htp]
\centering
\includegraphics[width=\linewidth]{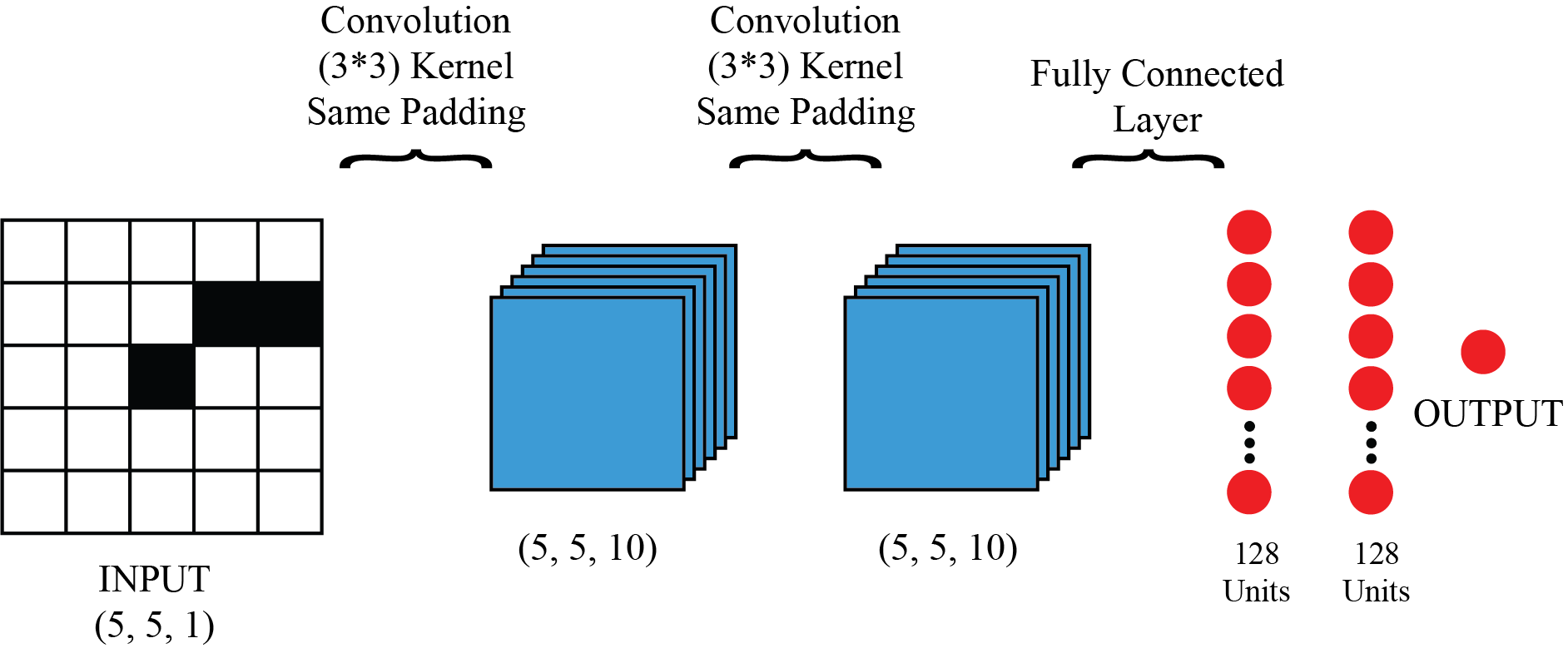}
\caption{Critic network architecture}
\label{fig4}
\end{figure}

In this paper, the authors have employed two simple CNNs to learn the policy and the expected cost at each observation, represented by a $5\times 5$ grid occupied by obstacles. Fig.~\ref{fig3} depicts the schematic of the architecture used as actor. The output of the actor network is a $5 \times 5$ matrix. Each column of the output matrix is converted to a probability distribution via a softmax function. The selected action at each column is sampled from this probability distribution. It is noted that the distribution obtained from the first column of the output is irrelevant since the first column corresponds to the predetermined second point of the B-spline. The critic network then takes the $5 \times 5$ matrix as input and estimates the baseline (i.e., the expected cost) to calculate the advantage function. The configuration of the critic network is represented in Fig.~\ref{fig4}. The loss function used to update the weights of the critic network is Mean Squared Error function:
\begin{equation}
\text{MSE}(C, \hat{C}) = \frac{1}{l} \sum_{i=1}^{l} (C_i - \hat{C}_i)^2
\end{equation}
where $l$ is the batch size, and $\hat{C}$ is the estimated cost.


\subsection{Target Generalization}
\begin{figure}[htp]
\centering
\includegraphics[width=0.5\linewidth]{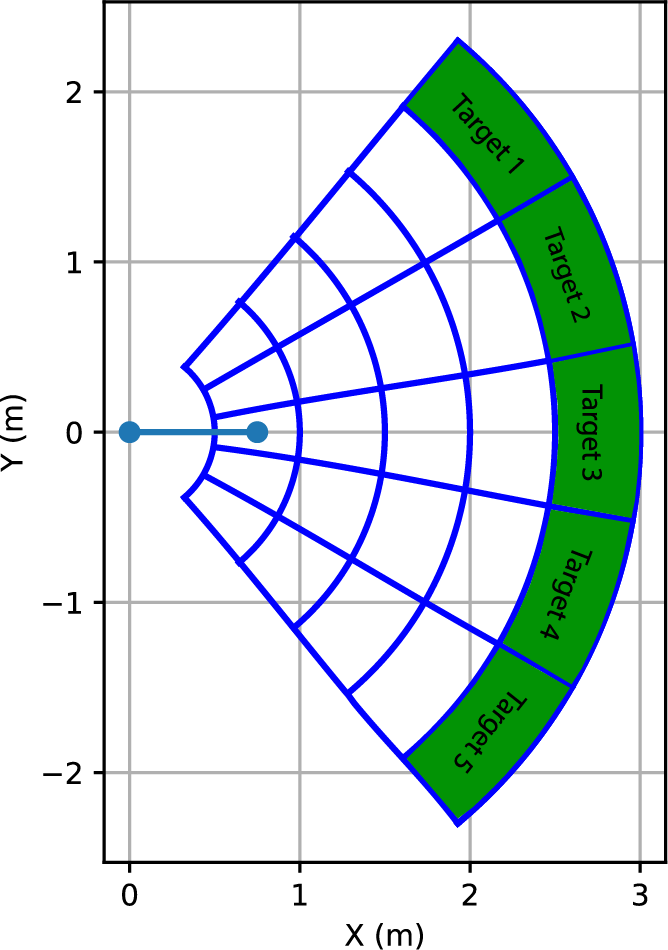}
\caption{Five potential temporary goal locations}
\label{fig5}
\end{figure}

One of the primary challenges encountered when applying CNN-based methods for path planning tasks is the incorporation of the target location into the network. Due to the robotic vehicle's limited field of view, achieving a globally optimized solution is not feasible. However, akin to how humans confront this dilemma in real-life scenarios, the robot is tasked with selecting a goal location within its field of view. In this approach, the distance between the final target and all five potential final points of the B-spline are calculated. Subsequently, we select the point closest to the final target, referred to as the "normalized target" in this study (Fig.~\ref{fig5}). The set containing all five normalized targets is:
\begin{equation*}
    \bm{G} = \{\mathcal{G}_1, \mathcal{G}_2, ..., \mathcal{G}_5\}
\end{equation*}
and the final target is denoted as $\mathcal{G}_{\text{final}}$.

SmartBSP trains five distinct actor networks, each targeting a different potential normalized target. In other words, the $C_{\text{dist}}$ component of (\ref{eq1}) represents the distance between the final point of the path and the normalized target. Subsequently, the network associated with the normalized target suggests a resulting optimized path.

However, when the path to the normalized target location is obstructed, the actor fails to propose a collision-free path. In such instances, it is assumed that the robot has encountered either a wall or a non-convex obstacle. The proposed method does not offer a path for all non-convex obstacles. To address such scenarios, however, the normalized target is shifted to that with the minimum cost. Cost $C_i(\mathcal{S}, \mathcal{A})$ is calculated by
\begin{equation}
C_i(\mathcal{S}, \mathcal{A}) = \rho_2 C_{\text{curv}} + \rho_3 C_{\text{obs}}
\end{equation}

This temporary target is denoted as $\mathcal{G}_{\text{temp}}$. Here, $C_i, i = 1, \ldots, n$, represents the cost associated with normalized target $i$ as shown in Fig.~\ref{fig5}. The set of control points associated with $\mathcal{G}_{\text{temp}}$, is denoted by $\mathcal{A}_{\text{temp}}$. This adjustment (i.e., incorporating $\mathcal{G}_{\text{temp}}$ and $\mathcal{A}_{\text{temp}}$) continues until the normalized target is no longer obstructed. The SmartBSP method for obstructed paths is summarized in Algorithm~\ref{alg:path_following}. Situations where even this adjustment of the goal fails can be categorized into two different types: (1) when all the temporary targets are obstructed (a rare occurrence, particularly when the path update rate is high or the field of view is wide) and (2) when extreme non-convexities are preset, such as U-shaped obstacles. Here, the authors assume that neither of these situations occur.
\begin{table}[htbp]
\caption{Training parameters}
\begin{center}
\begin{tabular}{lr}
\hline
\textbf{Parameter} & \textbf{Value} \\
\hline
batch size & $10$ \\
\# of epochs & $3$ \\
learning rate & $0.001$ \\
$\epsilon$ (PPO clipping constant) & $0.2$ \\
\# of updates per iteration & 5 \\
\hline
\end{tabular}
\label{tab1}
\end{center}
\end{table}
\section{Simulation Results}
To train each of the five aforementioned networks, 10,000 $5 \times 5$ grid samples with randomly distributed obstacles are generated. The training hyperparameters are presented in Table~\ref{tab1}.

Table~\ref{tab2} represents the success rates of the trained networks. Here, success is defined as the ability to avoid all obstacles present in the local grid.
\begin{table}[htbp]
\caption{Success rate of Trained networks}
\begin{center}
\begin{tabular}{cc}
\hline
\textbf{Trained Neural Networks} & \textbf{Success Rate} \\
\hline
NN 1 & $96\%$ \\
NN 2 & $95\%$ \\
NN 3 & $95\%$ \\
NN 4 & $95\%$ \\
NN 5 & $97\%$ \\
\hline
\end{tabular}
\label{tab2}
\end{center}
\end{table}

Since grids are occupied by ones and zeros, causing symmetries to occur in many samples, choosing the appropriate seed plays an important role in the success rate of trained networks.
\begin{algorithm}[H]
\caption{SmartBSP Algorithm}
\label{alg:path_following}
\begin{algorithmic}
\While{target position is not reached}
    \State Acquire sensor data $\mathcal{S}$ using the LIDAR sensor
    \State Update the local occupancy grid with observed points
    \If{$\mathcal{S} \neq \emptyset$}
        \State Calculate the coordinates of the normalized targets
        \State Choose the network associated with the closest target
        \State Generate $\mathcal{A}$ $\Rightarrow$ Calculate $C(\mathcal{S},\mathcal{A})$
        \If{$C_{\text{obs}}(\mathcal{S},\mathcal{A})=1$}
            \begin{align*}
                \qquad \mathcal{G}_{\text{temp}} = \mathcal{G}_i \  \text{where} \ i = \argmin_{i \in \{1,2,...,n\}} C_i(\mathcal{S},\mathcal{A})
            \end{align*}
            \State Generate $\mathcal{A}_{\text{temp}}$
        \EndIf    
    \Else
    \State $\mathcal{G} = \mathcal{G}_{\text{final}}$
    \EndIf
    \State Follow 10\% of the planned path segment.
\EndWhile
\end{algorithmic}
\end{algorithm}

The steps followed in simulation and experimental sections are summarized in Algorithm~\ref{alg:path_following}. Depending on the complexity of the environment, one can adjust the path update rate accordingly. In environments with numerous obstacles, the path should be updated more frequently. In this study, the path is updated when $10\%$ of the path is traversed. The chosen vehicle for the simulation is a simple differential drive robot which utilizes a PID controller to follow the path. The robot and the sensor parameters are represented in Table~\ref{tab3}.  The PID controller coefficients are adjusted to minimize tracking error. Since the controller can significantly affect the path planner performance, using a well-designed controller is vital.
\begin{table}[htbp]
\caption{The differential drive robot and range sensor parameters}
\begin{center}
\begin{tabular}{lr}
\hline
\textbf{Parameter} & \textbf{Value} \\
\hline
Wheelbase & $0.15~m$ \\
Wheel Radius & $0.05~m$ \\
Sensor Range & $3~m$ \\
Field of View & $100~deg$ \\
Radial Intervals & $0.5~m$ \\
Angular Intervals & $20~deg$ \\
\hline
\end{tabular}
\label{tab3}
\end{center}
\end{table}

To demonstrate the capabilities of the proposed algorithm, various scenarios are tested against the path planner. The obstacles' edges in each scenario are represented by points to simulate feedback from conventional range sensors. Notably, the environment itself is not divided into grid cells. When the vehicle seeks to update its path at any given time step, it observes the circular segment ahead of it. Each circular grid cell containing a certain prescribed number of points detected by LiDAR sensor is then marked as an obstacle. The output of this process is depicted in Fig.~\ref{fig1}(b).

\subsection{Scenario 1: Environments with multiple small obstacles}
\begin{figure}[htp]
\centering
\includegraphics[width=1\linewidth]{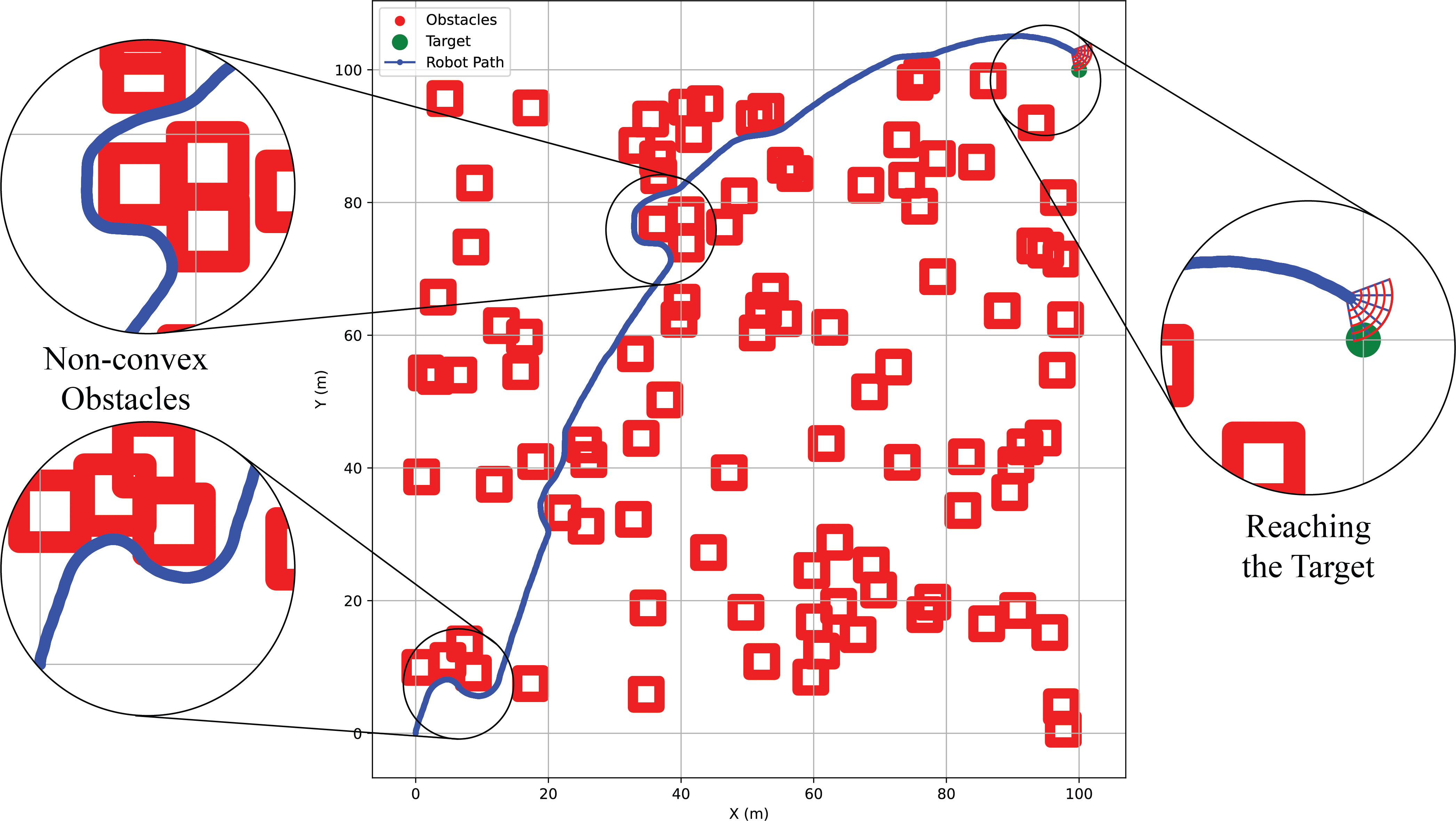}
\caption{Path planning result in scenario 1}
\label{fig6}
\end{figure}

In this scenario, the environment is populated with obstacles of uniform size, randomly positioned throughout the search area. The vehicle navigates from the origin towards the target, piece by piece, relying on its limited field of view. The simulation concludes when the robot's distance to the target is less than a chosen threshold of $3~m$. Fig.~\ref{fig6} illustrates how the robot can overcome non-convex obstacles. The level of non-convexity significantly impacts the algorithm's performance. Specifically, if addressing non-convex situations requires information on past observed environments and decisions, the algorithm may fail. Such environments are not considered in this study.

\subsection{Scenario 2: Passing large obstacles}
\begin{figure}[htp]
\centering
\includegraphics[width=1\linewidth]{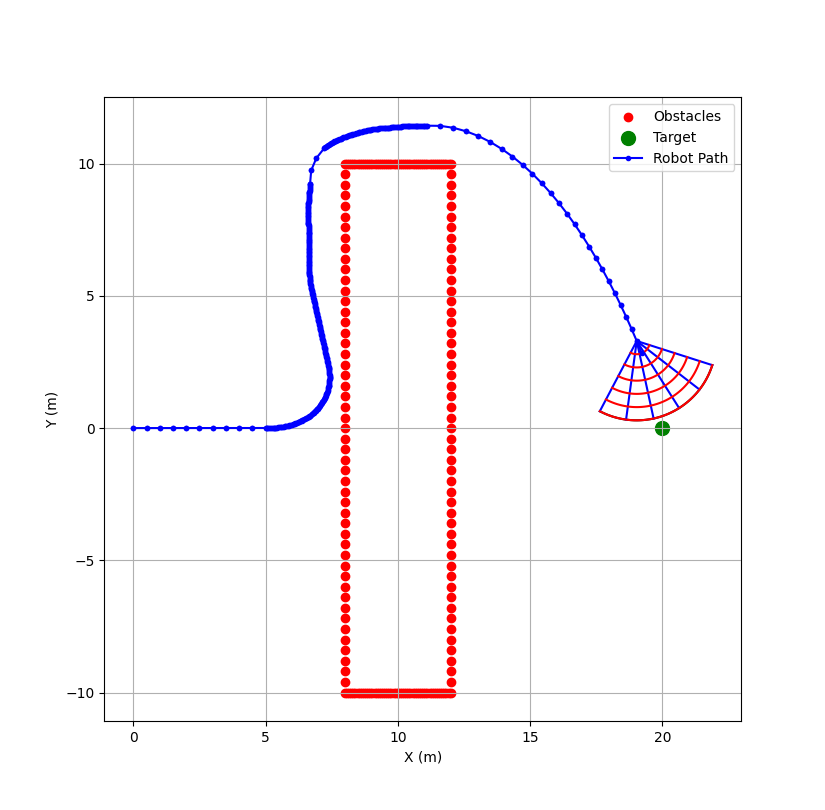}
\caption{Path planning result in scenario 2}
\label{fig7}
\end{figure}

Passing a large obstacle without knowledge of its endpoint poses a significant challenge. When the vehicle encounters the wall, it must decide whether to turn right or left. Given the limited field of view, making an optimal decision regarding the distance traveled by the vehicle becomes impossible. Here, the assumption is made that the wall is short, meaning that turning right or left would not significantly impact the solution. Fig.~\ref{fig7} illustrates the performance of the algorithm in such scenarios. 

\section{Experimental Results}
\begin{figure}[htp]
\centering
\includegraphics[width=1\linewidth]{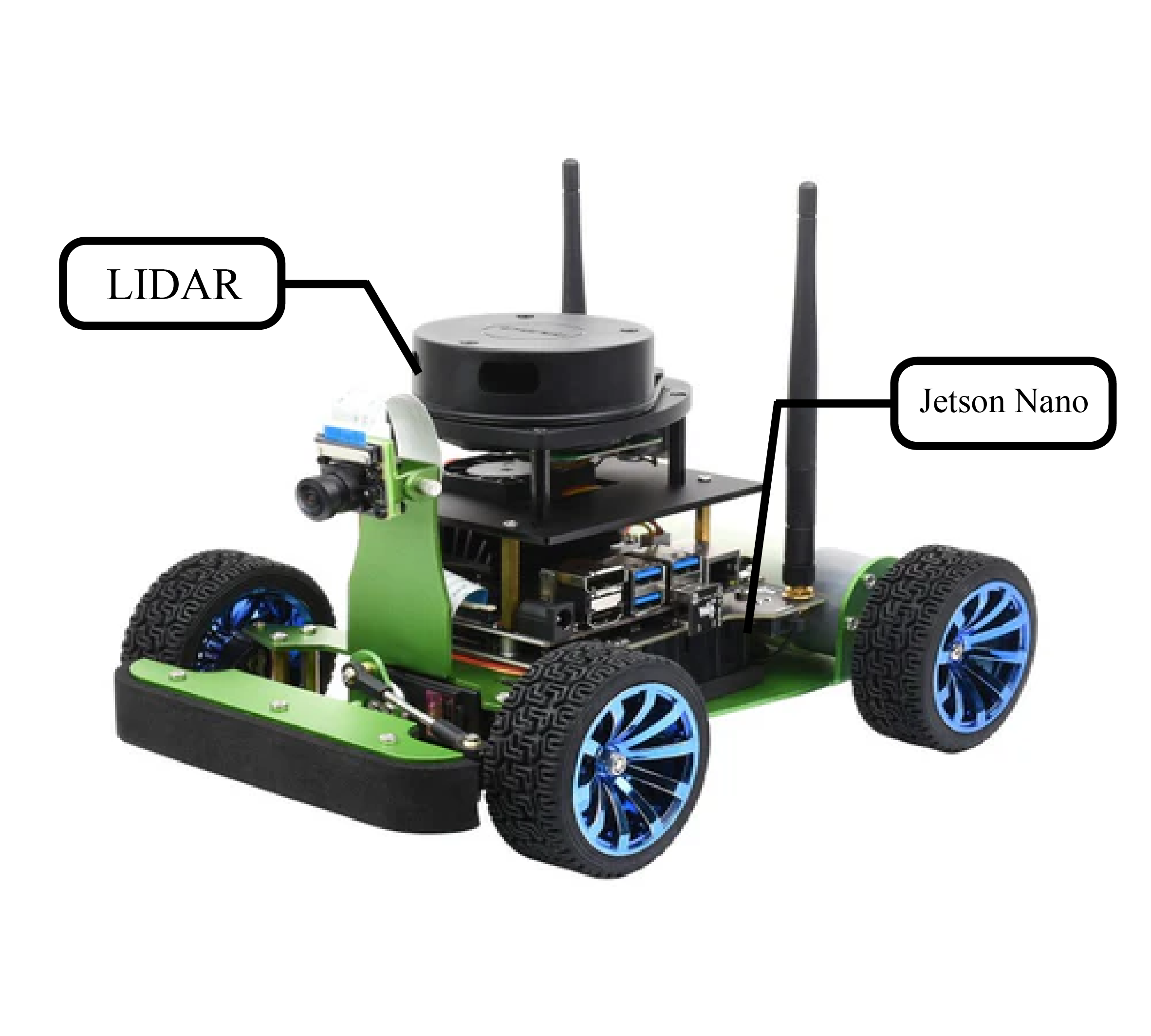}
\caption{Jetracer platform}
\label{fig8}
\end{figure}

To test the algorithm performance in real situations, the Jetracer platform is employed. The JetRacer robot, shown in Fig.~\ref{fig8}, is a small-scale, autonomous vehicle designed for experimentation and research purposes. Equipped with a variety of sensors and a Jetson Nano onboard computer, the JetRacer is capable of executing complex navigation tasks in dynamic environments.

The JetRacer platform is equipped with a $360^\circ$ LiDAR sensor to detect obstacles. The LiDAR range is $12m$. To mimic the algorithm used in this paper, however, the LiDAR output is filtered to limit the range and the field of view to match those used in simulations. The selection of these parameters, such as the range and field of view of the LiDAR sensor, depends on the maneuverability of the robot and the number and shape of obstacles present in the environment.

\begin{figure}[htp]
\centering
\includegraphics[width=1\linewidth]{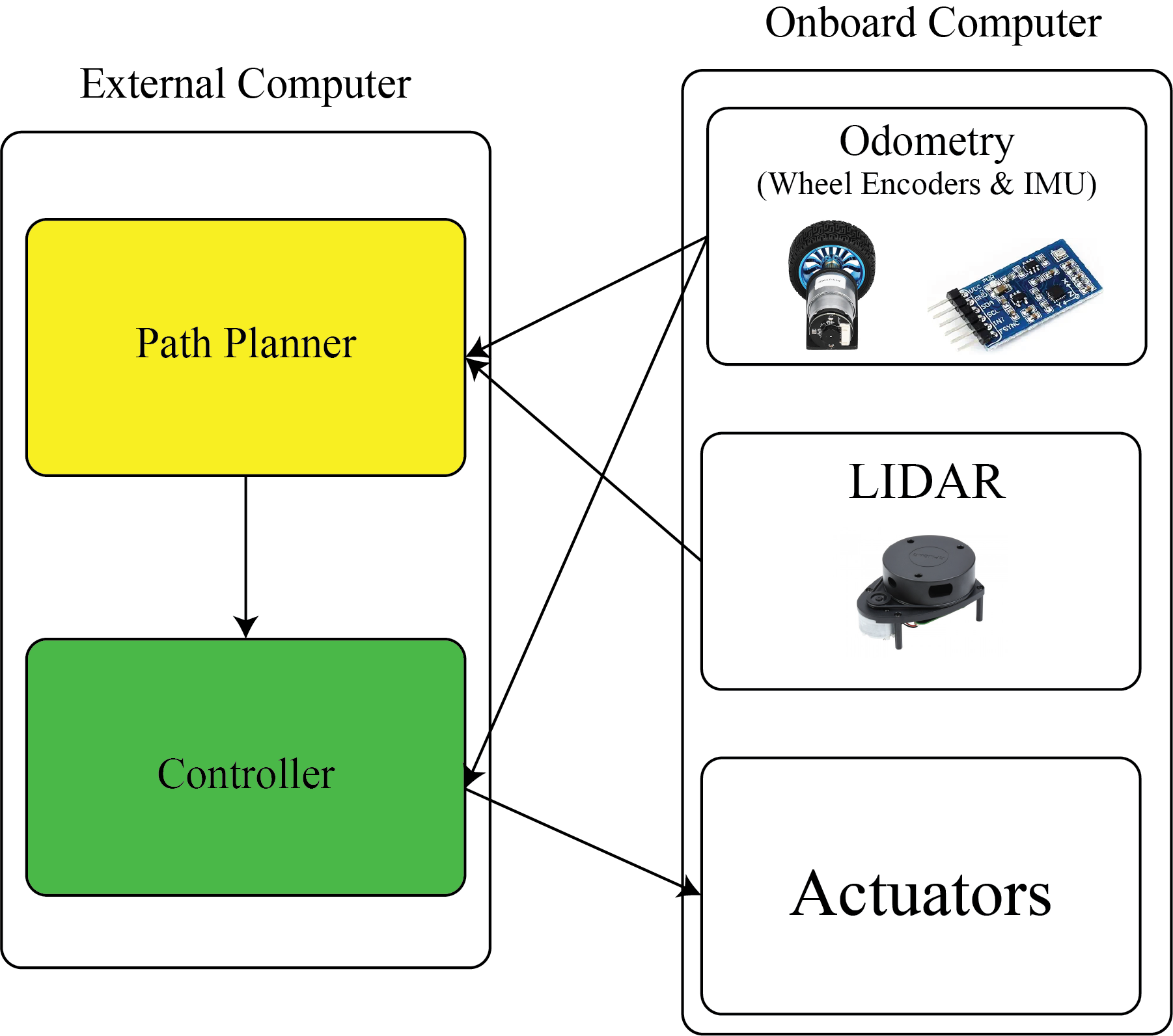}
\caption{ROS modules configuration}
\label{fig9}
\end{figure}

To maintain the desired path, a straightforward PID controller is utilized. The vehicle maintains a constant forward speed, while the controller dynamically adjusts the steering angle based on the next waypoint provided by the path planner. Fig.~\ref{fig9} illustrates the interconnectedness of various components of the platform through the Robot Operating System (ROS). The entire system is designed with modularity in mind, facilitating easy substitutions of various controllers and path planners.

\begin{figure}[htp]
\centering
\includegraphics[width=1\linewidth]{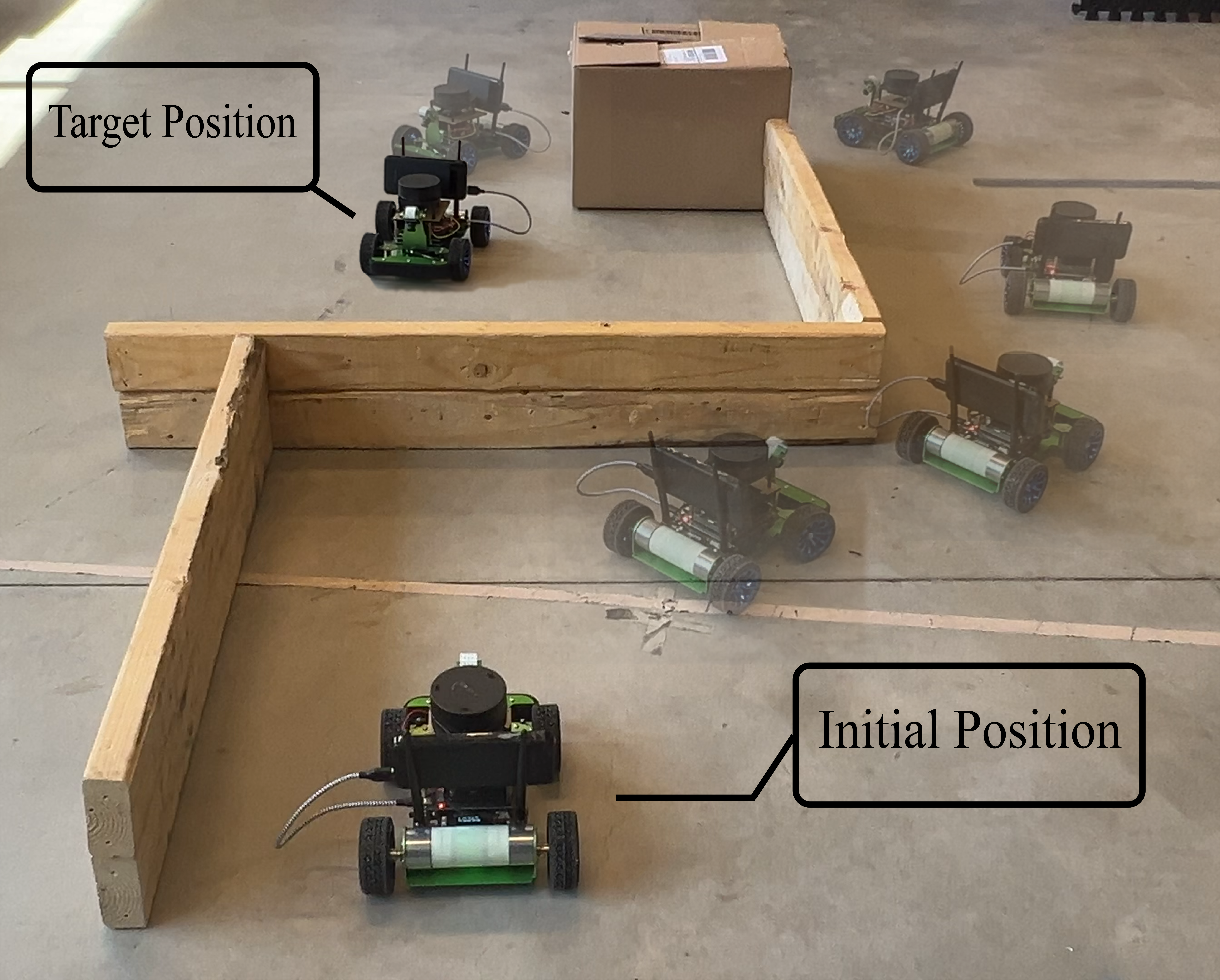}
\caption{Test 1: Reaching Target Located Behind a Wall-Shaped Obstacle}
\label{fig10}
\end{figure}

\begin{figure}[htp]
\centering
\includegraphics[width=1\linewidth]{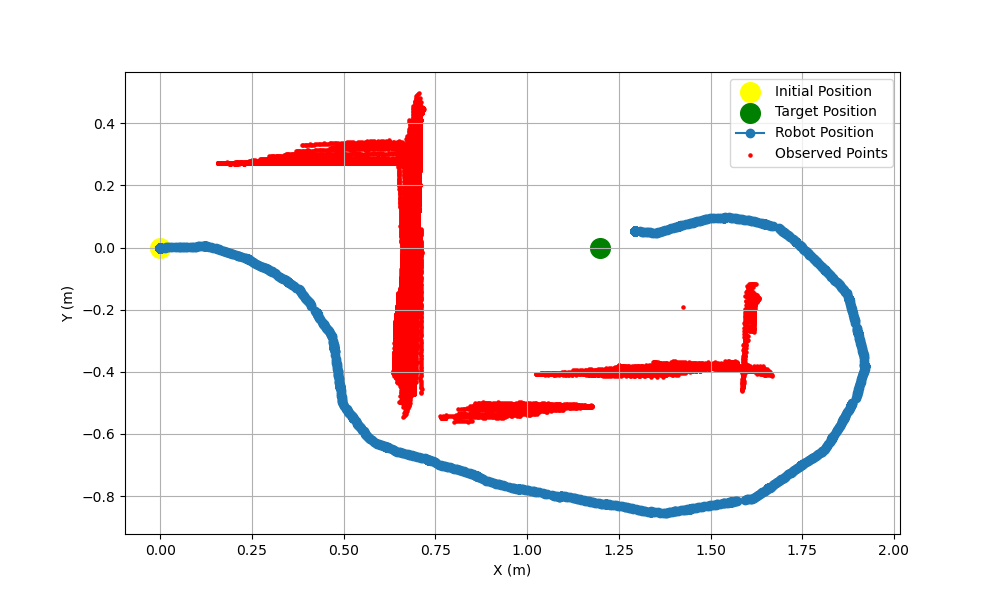}
\caption{Test 1: Robot Path and Observed Environment}
\label{fig11}
\end{figure}

To assess the algorithm, the vehicle is tested in two different scenarios. Fig.~\ref{fig10} and Fig.~\ref{fig11} present the results of the first test. The final goal is located behind a wall, requiring the robot to circumvent a non-convex wall to reach it. To perform this test, the vehicle's field of view is set to $120^\circ$ and the sensor range to $0.8\,\text{m}$. For both experimental tests, the robot has no prior knowledge of the environment. The starting position, denoted by yellow, is considered the origin. The robot is programmed to stop when it is within a predefined threshold distance from the target position. The red points represent the observations of obstacles from the LiDAR sensor. The entire path is planned incrementally, utilizing the robot's limited field of view.

\begin{figure}[htp]
\centering
\includegraphics[width=1\linewidth]{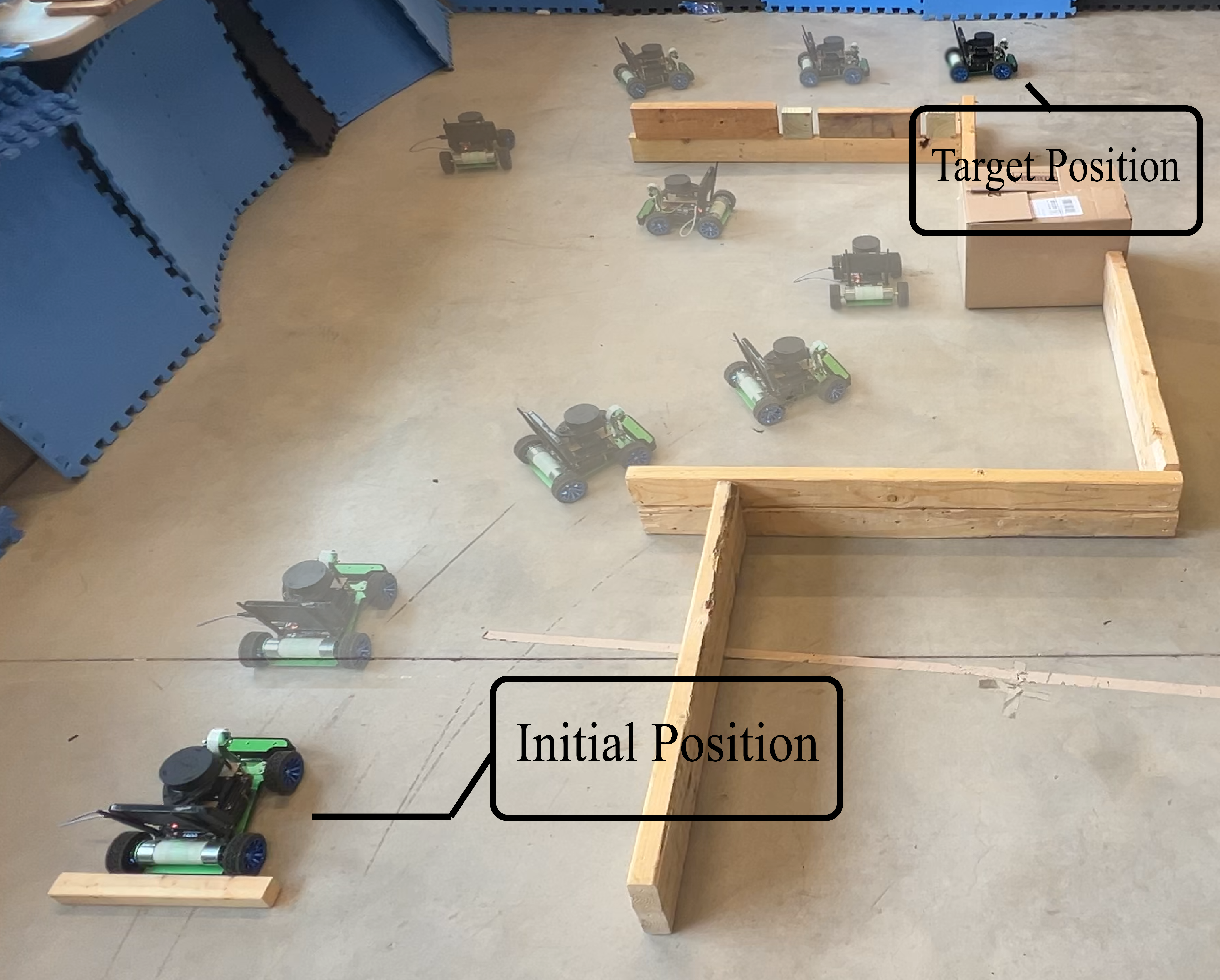}
\caption{Test 2: Avoiding non-convex obstacle}
\label{fig12}
\end{figure}

\begin{figure}[htp]
\centering
\includegraphics[width=1\linewidth]{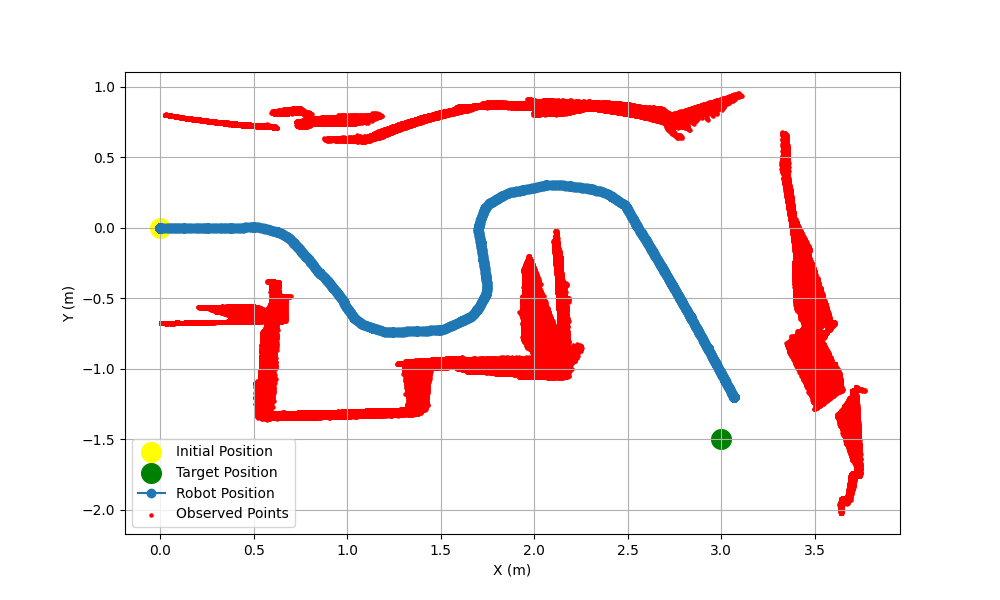}
\caption{Test 2: Robot Path and Observed Environment}
\label{fig13}
\end{figure}

In the second test, the robot is subjected to a more complex situation. Fig.~\ref{fig12} and Fig.~\ref{fig13} illustrate the results of the second test. During the test, it is observed that the field of view plays a crucial role in the vehicle's success rate. To ensure adequate performance, the robot requires a field of view of $180^\circ$. Here, changing the field of view is achieved simply by adjusting the filter on the LiDAR sensor, without the need to retrain the networks. In other words, the proposed method is generalizable to different environments by appropriately selecting the sensor range and field of view. 

\section{Conclusion}
In this paper, the authors introduce SmartBSP, an AI-based path planning approach designed for autonomous robotics in complex and unknown environments. By leveraging Proximal Policy Optimization (PPO) combined with Convolutional Neural Networks (CNNs) and an Actor-Critic architecture, the proposed method effectively processes limited LIDAR inputs to generate optimal paths in real-time. The integration of B-spline curves and a nuanced cost function accounting for path curvature, endpoint proximity, and obstacle avoidance further enhances the robustness and adaptability of the system.

Through extensive analytical simulations and lab experiments, the authors demonstrate that the proposed method maintains good performance in spite of challenging scenarios, including environments with multiple small obstacles and large obstacles. Experimental results show that adjusting the field of view and sensor range significantly impacts the success rate of the vehicle, highlighting the importance of these parameters in path planning tasks. Moreover, the flexibility of the SmartBSP approach allows for easy adaptation to different environments without the need for retraining the networks, showcasing its potential for practical applications in diverse operational scenarios.

Future work will focus on further refining the algorithm to take into account more complex non-convex obstacles and exploring the integration of additional sensor modalities to enhance the perception capabilities of the autonomous system. Additionally, the authors aim to apply the SmartBSP framework to other types of autonomous vehicles and investigate its performance in large-scale outdoor environments.

\end{document}